\documentclass[lettersize,journal]{IEEEtran}
\usepackage{amsmath,amsfonts}
\usepackage{algorithmic}
\usepackage{algorithm}
\usepackage{array}
\usepackage[caption=false,font=normalsize,labelfont=sf,textfont=sf]{subfig}
\usepackage{textcomp}
\usepackage{stfloats}
\usepackage{url}
\usepackage{verbatim}
\usepackage{graphics} 
\usepackage{graphicx}
\usepackage{booktabs}
\usepackage{pifont}
\usepackage{amsfonts}
\usepackage{amsmath}
\usepackage{times}
\usepackage{float}
\usepackage{multicol}
\usepackage{multirow}
\usepackage{array}
\usepackage{booktabs}
\usepackage[dvipsnames]{xcolor}
\usepackage{color, colortbl}
\usepackage{soul}
\usepackage{caption}
\usepackage{stmaryrd}
\usepackage{bm}
\newcommand{\Loss}{\mathcal{L}}

\definecolor{LightCyan}{rgb}{0.95,0.95,1.0}
\definecolor{LightYellow}{rgb}{0.98,0.99,0.84}

\begin{document}
\title{SelfOccFlow: Towards end-to-end self-supervised \\ 3D Occupancy Flow prediction}

\author{Xavier Timoneda$^{1}$, Markus Herb$^{1}$, Fabian Duerr$^{1}$, Daniel Goehring$^{2}$

\thanks{© 2026 IEEE.  Personal use of this material is permitted.  Permission from IEEE must be obtained for all other uses, in any current or future media, including reprinting/republishing this material for advertising or promotional purposes, creating new collective works, for resale or redistribution to servers or lists, or reuse of any copyrighted component of this work in other works.}
\thanks{$^{1}$Xavier Timoneda, Markus Herb, and Fabian Duerr are with the Onboard Fusion team at CARIAD SE, Volkswagen Group, Ingolstadt, Germany. \\
{\tt\small xavier.timoneda.comas@cariad.technology}}%
\thanks{$^{2}$Daniel Goehring is with the Dahlem Center for Machine Learning and Robotics group at Freie Universit{\"a}t Berlin, Germany.}}

\maketitle

\begin{abstract}

Estimating 3D occupancy and motion at the vehicle's surroundings is essential for autonomous driving, enabling situational awareness in dynamic environments. Existing approaches jointly learn geometry and motion but rely on expensive 3D occupancy and flow annotations, velocity labels from bounding boxes, or pretrained optical flow models. We propose a self-supervised method for 3D occupancy flow estimation that eliminates the need for human-produced annotations or external flow supervision. Our method disentangles the scene into separate static and dynamic signed distance fields and learns motion implicitly through temporal aggregation. Additionally, we introduce a strong self-supervised flow cue derived from features' cosine similarities. 
We demonstrate the efficacy of our 3D occupancy flow method on SemanticKITTI, KITTI-MOT, and nuScenes.

\end{abstract}


\section{Introduction}

3D occupancy prediction \cite{RenderOcc, Surroundocc, OccNeRF, SelfOcc} is a fundamental task in computer vision and autonomous driving, aimed at estimating which regions around a vehicle are occupied or free based on current (and past) sensor data. To enable real-time predictions, occupancy models limit their scope within a bounding box surrounding the ego vehicle and discretize it into equally sized voxels \cite{SemanticKITTI, nuScenes, openoccv2}. These models use data from sensors typically present in modern cars, such as cameras, LiDAR and Radar. Camera-based approaches, like ours, rely only on camera inputs to infer occupancy.

Early camera-based approaches \cite{LMSCNet, MonoScene, Yan2021} are trained with costly 3D occupancy annotations, which require extensive expert 3D labeling. Inspired by neural fields \cite{Mescheder2019}, more recent methods \cite{BTS, SceneRF, OccFlowNet, SelfOcc} implicitly model geometries as a continuous field, enabling ray marching-based supervision. During training, the continuous field is queried at arbitrary locations along LiDAR and camera rays from the current or adjacent timesteps to supervise depths using LiDAR ranges or photometric reprojection losses \cite{SelfOcc, LetOccFlow}. Following NeuS' \cite{NeuS} formulation, SelfOcc \cite{SelfOcc} models geometry as a Signed Distance Field (SDF). An SDF represents the signed Euclidean distance from any point to its nearest surface, being positive at empty regions, negative inside objects, and zero at surfaces. The gradient of the SDF is a unit-norm vector pointing to the closest surface (surface normal) \cite{NeuS}, which provides rich contextual geometric information. Like \cite{SelfOcc} and \cite{LetOccFlow}, we adopt SDFs to represent our model's predicted geometry.

Supervising occupancy models with adjacent frames is crucial for methods based on multi-view consistency \cite{SelfOcc, LetOccFlow} and improves temporal consistency for LiDAR-supervised methods \cite{OccFlowNet}. However, dynamic objects can introduce inconsistencies, which some works mitigate by using 3D bounding boxes and their velocity labels \cite{OccFlowNet}, while others \cite{SelfOcc} ignore dynamic or occluded pixels using min-loss and auto-masking strategies from Monodepth \cite{monodepth2}.

In dynamic driving scenes, geometry estimation alone may not suffice for a comprehensive understanding of the scene. Recent 3D occupancy flow models jointly predict occupancy and scene flow \cite{openoccv2, ViewFormer, LetOccFlow}, estimating both occupancy status and short-term motion vectors for every point. These models rely on current and past-frame inputs to capture temporal context, enabling motion estimation and improving cross-frame geometric consistency \cite{LetOccFlow}. However, obtaining 3D flow annotations is complex and expensive \cite{emernerf}, which limits the scalability of 3D occupancy flow models. LetOccFlow \cite{LetOccFlow} addresses this issue using photometric self-supervision, but still relies on pseudo-labels from pretrained 2D optical flow models \cite{unimatch, cotracker}, which are trained on standard datasets using optical flow annotations and fine-tuned for each target dataset.

\begin{figure}[t]
\centering
\includegraphics[width=0.485\textwidth]{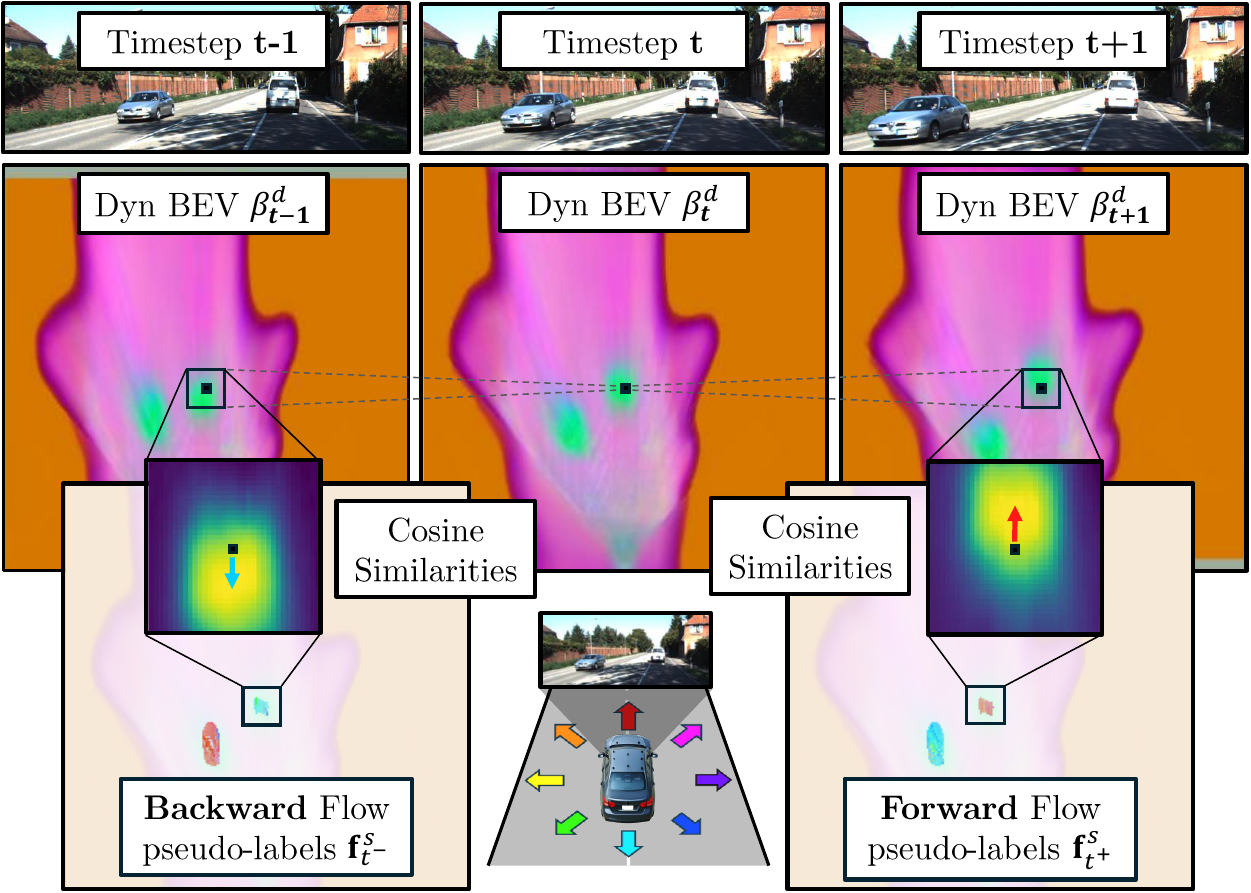}
\caption{\textbf{Similarity flow self-supervision on SemanticKITTI}. We provide explicit flow supervision by comparing the dynamic BEV features at current time $t$ and its adjacent frames $t\!\pm\!1$ aligned by ego motion. The flow pseudo-labels $\mathbf{f}^{s}_{t^-}, \mathbf{f}^{s}_{t^+}$ are obtained from the cosine similarities between each cell in the current timestep and its $N\!\times\!N$ neighbors at $t\!\pm\!1$.}
\label{fig:bevfeatssim}
\end{figure}

This work presents the first 3D occupancy flow estimation method that jointly learns geometry and motion without using occupancy labels, flow annotations, or pretrained optical flow networks. Instead, it relies solely on spatio-temporal consistency and foundation model-guided self-supervision. 

Our main contributions are:
\begin{itemize}
    \item A 3D occupancy flow model that explicitly disentangles static and dynamic SDFs, enabling learning geometry in occluded regions using static rays from neighbor frames.
    
    \item Per-field temporal aggregation mechanisms with flow warping on the dynamic field, enhancing cross-frame consistency and enabling implicit flow learning.

    \item A self-supervised similarity flow loss computed using the cosine similarities of the dynamic features (Fig. \ref{fig:bevfeatssim}).

\end{itemize}

\begin{figure*}[h!]
\centering
\includegraphics[width=0.99\textwidth]{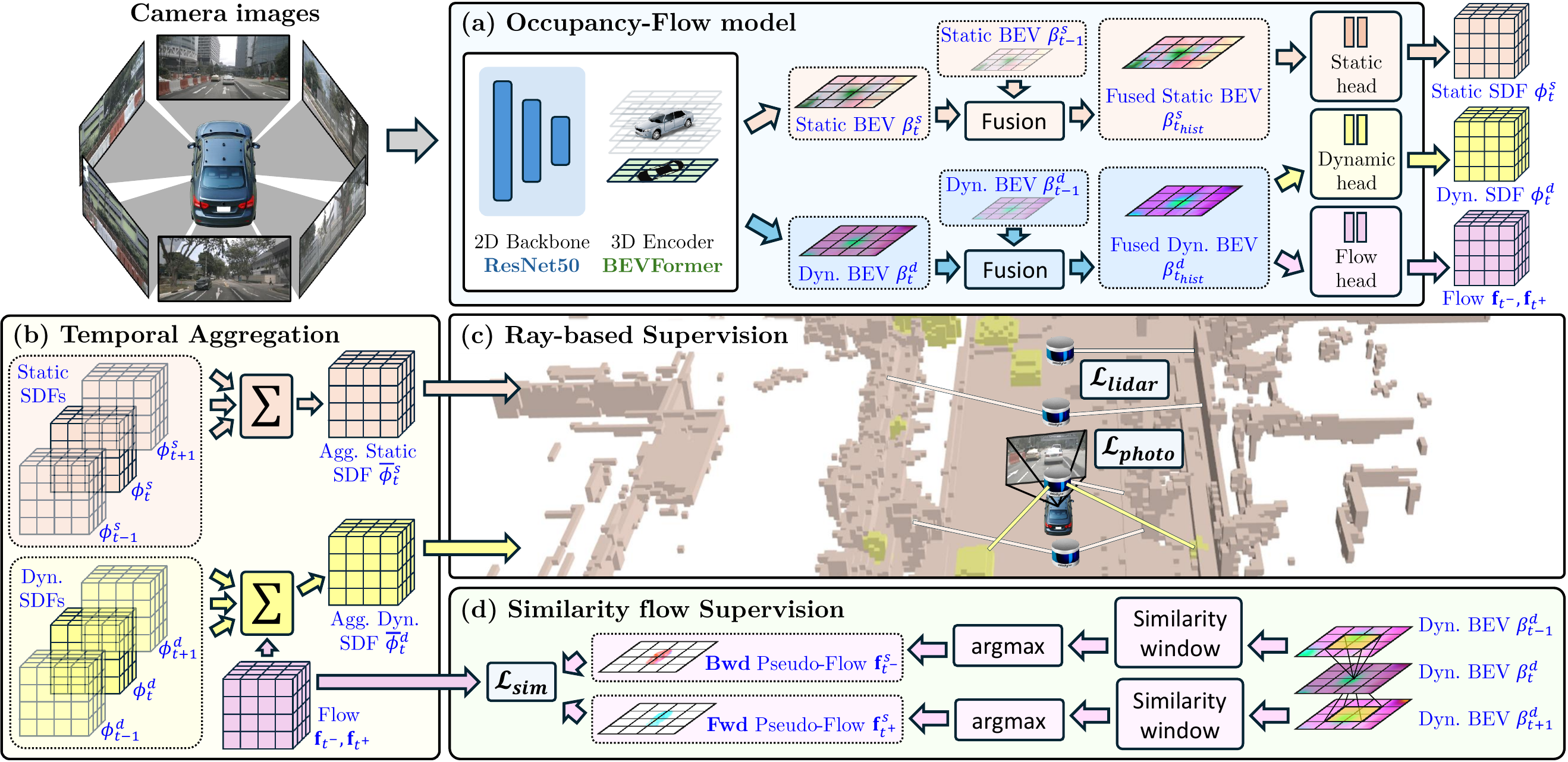}
\caption{\textbf{Overview of our method.} 
\textbf{a)} \textbf{Occupancy-Flow model.} Baseline pipeline used for both training and inference. The multi-view images from time $t$ are processed by a 2D backbone, and their features are fused in the 3D encoder. Half of the encoder's attention heads produce static BEV features $\beta^{s}_{t}$, and the other half produce dynamic ones $\beta^{d}_{t}$. Both $\beta^{s}_{t}$ and $\beta^{d}_{t}$ are fused with their previous frame features $\beta^{s}_{t-1}$, $\beta^{d}_{t-1}$ aligned by ego motion. The fused static features $\beta^{s}_{t_{\text{hist}}}$ are used to predict the static SDF $\phi^{s}_{t}$, while the fused dynamic features $\beta^{d}_{t_{\text{hist}}}$ are used to predict the dynamic SDF $\phi^{d}_{t}$, and flow $\mathbf{f}$.
\textbf{b)} \textbf{Temporal aggregation.} Static field predictions $\phi^{s}_{t}$ are aggregated with $\phi^{s}_{t-1}$ and $\phi^{s}_{t+1}$, aligned by ego motion. For the dynamic field $\phi^{d}_{t}$, the sampling positions at $\phi^{d}_{t-1}$ and $\phi^{d}_{t+1}$ are first warped by the flow $\mathbf{f}$, enabling implicit flow learning. 
\textbf{c)} \textbf{Ray-based supervision} of aggregated fields $\bar{\phi}^{s}_t, \bar{\phi}^{d}_t$ using photometric $\Loss_{photo}$ and LiDAR $\Loss_{lidar}$ losses. $\Loss_{photo}$ is applied to the blended field $\phi^{b}_{t}$ using camera rays from time $t$ only. $\Loss_{lidar}$ supervises dynamic predictions $\phi^{d}_{t}$ using dynamic LiDAR rays from $t$, and static predictions $\phi^{s}_{t}$ using static LiDAR rays from $t \pm k$ neighbors, leveraging its stationary nature.
\textbf{d)} \textbf{Similarity flow supervision} using auto-generated flow pseudo-labels $\mathbf{f}^{s}_{t^-}, \mathbf{f}^{s}_{t^+}$, obtained as the $\text{argmax}$ of the cosine similarity between $\beta^{d}_{t}$ and $\beta^{d}_{t\pm1}$ over $N\!\times\!N$ neighboring cells.}
\label{fig:overallpipeline}
\end{figure*} 

\section{Related work}
\textbf{3D occupancy prediction.} This task \cite{TPVFormer, openoccv2} has emerged as an alternative to traditional object-centric pipelines for autonomous driving, providing fine-grained spatial understanding of the vehicle’s surroundings.
Early works predict 3D occupancy from explicit 3D inputs such as depth \cite{Garbade2019, Li2020, Liu2018}, occupancy grids \cite{Garbade2019, LMSCNet, Yan2021}, pointclouds \cite{Rist2021, Zhong2020}, or truncated signed distance functions \cite{Chen2020, Song2017, Zhang2018}. Recent camera-based methods \cite{TPVFormer, MonoScene} predict semantic occupancy from only images, but rely on expensive 3D annotations. Concurrently, self-supervised learning methods have been developed across many machine learning applications \cite{Doersch2015, SimCLR, Timoneda2024} to reduce label reliance. They leverage inherent structures and relationships in the data to create meaningful training signals without human intervention. Some recent works learn 3D occupancy in a self-supervised manner for monocular scenarios \cite{BTS, SceneRF} and surround views \cite{SelfOcc, LetOccFlow}. However, most of them operate only on static environments or design losses that ignore moving objects when enforcing temporal consistency. Fewer works jointly learn geometry and motion in dynamic scenes. The work \cite{LetOccFlow} jointly learns occupancy and scene flow using pretrained 2D optical flow models, while \cite{OccFlowNet} relies on velocity labels. Although effective, both methods rely on external supervision to deal with moving objects.

\textbf{Neural fields.} A central paradigm for 3D scene representation is neural fields \cite{Mescheder2019, DeepSDF}. This class of models introduces implicit representations as inductive biases to enable 3D understanding. Unlike mesh or voxel-based methods, neural fields represent geometry with continuous implicit functions parametrized by neural networks that map spatial coordinates to physical quantities such as color, density, occupancy, or signed distance. The most influential example is Neural Radiance Fields (NeRFs) \cite{NeRFs}, in which a Multi-Layer Perceptron (MLP) learns a continuous volumetric scene representation by mapping 3D coordinates and viewing directions into color and density. To synthesize a novel view of the scene, the MLP is queried at spatial locations along camera rays, and volume rendering integrates radiance over density-induced weights. Extensions such as NeuS \cite{NeuS} introduce signed distance functions for improved surface reconstruction, while dynamic NeRFs \cite{DNeRF, Li2024, emernerf} model time-varying scenes by adding a time variable to the input.

\section{Method}

This section establishes our problem mathematically and provides a detailed description of each system component.

\textbf{Problem formulation.} We train a perception model that, given a set of video frames from multi-view cameras, predicts occupancy and scene flow within a 3D volume around the ego vehicle. The model learns scene geometry and motion by generating spatio-temporal supervision cues from camera and LiDAR rays, and leverages foundation model-guided self-supervision for scene disentanglement.

\textbf{Static - dynamic disentanglement.} Outdoor driving scenes contain both static and dynamic elements. Static geometry can be effectively learned through multi-view consistency across timesteps thanks to its stationary nature. However, dynamic objects (e.g., cars, pedestrians) can move and complicate this process. Following dynamic NeRFs \cite{emernerf}, we simplify the learning of motion and geometries by disentangling the scene into static and dynamic parts, predicting separate static $\phi^{s}$ and dynamic $\phi^{d}$ SDFs, and obtaining the total blended SDF $\phi^{b}$ as their minimum. To ensure differentiability, we leverage the property $\min(x,y) = -\max(-x, -y)$ to approximate it as:

\begin{equation}
\phi^{b} = \min(\phi^{s}, \phi^{d}) \approx - \text{softmax}(-\phi^{s} \!\cdot\! \tfrac{a}{\tau}, -\phi^{d} \!\cdot\! \tfrac{a}{\tau})
\label{eq:sdfblending}
\end{equation}
where $a$ is a learnable parameter and $\tau$ is the temperature scalar, and both control the sharpness of the approximation. 

Unlike dynamic NeRFs, which separate scenes based on instantaneous motion (e.g., treating parked cars as static), our approach disentangles static and dynamic elements based on semantic classes. We chose this separation because we observed that our model's encoder shows a strong semantic bias, mainly due to the 2D backbone being initialized with ImageNet weights. This bias promotes early-layer separation, stable training and faster convergence than motion-based disentanglement. Furthermore, motion-based disentanglement introduces temporal ambiguity when an object’s state changes between frames (e.g., a parked car begins to move), compromising our consistency-based learning objectives. In contrast, our semantic disentanglement remains invariant to the object's motion status, yielding more stable geometry predictions.

To enforce this separation, we supervise the static $\phi^{s}$ and dynamic $\phi^{d}$ fields with static $\mathbf{r}^{s}$ and dynamic $\mathbf{r}^{d}$ rays. We classify supervisory rays into static and dynamic by querying a generic image foundation model \cite{GroundedSAM} for prompts of dynamic classes (e.g., \textit{car}, \textit{pedestrian}), which predicts dynamic masks from every training image. We classify LiDAR rays by projecting their endpoints into the camera plane and assigning them static/dynamic pseudo-labels using the masks. To reduce the noise caused by objects' motion and parallax effect, we voxelize the dynamic points, apply 3D connected components, and retain only the largest cluster per mask, discarding outliers. The kept rays are labeled as dynamic, and unassigned ones are labeled as static.

\textbf{Baseline model pipeline.} Our occupancy flow prediction model is depicted in Fig. \ref{fig:overallpipeline}. It takes as input surround-view images at time $t$ and extracts multi-scale features using a ResNet50 encoder \cite{ResNet} and a feature pyramid network, following \cite{SelfOcc}. A BEVFormer \cite{BEVFormer} encoder aggregates them into a unified Bird's-Eye-View (BEV) representation, with half of the attention heads encoding static features $\beta^{s}$ and the other half encoding dynamic ones $\beta^{d}$. Both $\beta^{s},\beta^{d}$ are fused with ego-motion-aligned static and dynamic features at $t\!-\!1$ with channel-wise concatenation. The fused static features $\beta^{s}_{t_{\text{hist}}}$ are processed by the static head to predict a static SDF $\phi^{s}_{t}$, while dynamic ones $\beta^{d}_{t_{\text{hist}}}$ are processed by the dynamic and flow heads, predicting a dynamic SDF $\phi^{d}_{t}$ as well as backward $\mathbf{f}_{t^-}$ and forward $\mathbf{f}_{t^+}$ flows. Here, $\mathbf{f}_{t^-}\!:= \mathbf{f}_{t \shortrightarrow t-1}(\mathbf{x})$ represents the estimated motion of a point $\mathbf{x}$ from frame $t$ to the previous frame, and $\mathbf{f}_{t^+}\!:= \mathbf{f}_{t \shortrightarrow t+1}(\mathbf{x})$ the motion from $t$ to the next one. This baseline model constitutes the inference-time network.

\textbf{Temporal aggregation.} During training, we enforce temporal aggregation of field predictions across adjacent frames before applying geometric losses. At each iteration, we query the baseline model at timesteps $t\!-\!1$, $t$, and $t\!+\!1$ aligning all them to $t$ coordinates according to ego motion. The aggregated static SDF $\bar{\phi}^{s}_t$ at time $t$ for a location $\mathbf{x} \in \mathbb{R}^3$ is obtained as:

\begin{equation}
\label{staticsdfagg}
\bar{\phi}^{s}_t(\mathbf{x}) = \lambda_{\text{ag}} \cdot \frac{ \phi^{s}_{t-1}(\mathbf{x}) + \phi^{s}_{t+1}(\mathbf{x})}{2} + \left(1 - \lambda_{\text{ag}}\right) \cdot \phi^{s}_t(\mathbf{x})
\end{equation}

where $\lambda_{\text{ag}}$ is the aggregation ratio scalar. Because static elements are stationary, the aligned static fields can be aggregated directly without warping $\mathbf{x}$. This aggregation promotes inter-frame consistency in the model's predicted geometry. The aggregated dynamic SDF is obtained similarly as:

\begin{equation}
\label{dynsdfagg}
\bar{\phi}^{d}_t(\mathbf{x}) = \lambda^{d}_{\text{ag}}(\mathbf{x}) \cdot \frac{\phi^{d}_{t-1}(\mathbf{\hat{x}^{-}}) + \phi^{d}_{t+1}(\mathbf{\hat{x}^{+}})}{2}    + \left(1 - \lambda^{d}_{\text{ag}}(\mathbf{x})\right) \cdot \phi^{d}_t(\mathbf{x})
\end{equation}

where the aligned dynamic fields $\phi^{d}_{t-1}$ and $\phi^{d}_{t+1}$ are sampled at warped locations:

\begin{equation}
\mathbf{\hat{x}^{-}} = \mathbf{x} + \mathbf{f}_{t^-}(\mathbf{x}), \quad \mathbf{\hat{x}^{+}} = \mathbf{x} + \mathbf{f}_{t^+}(\mathbf{x})
\end{equation}

with $\mathbf{f}_{t^-}$ and $\mathbf{f}_{t^+}$ denoting the backward and forward flow predictions, respectively. The spatially varying blending weight is defined as $\lambda^{d}_{\text{ag}}(\mathbf{x}) = \lambda_{ag} \Phi^{d}_{a}(\mathbf{x})$, where $\Phi^{d}_{a}(\mathbf{x}) = (1+e^{a \cdot \phi^{d}_t(\mathbf{x})})^{-1}$ is the dynamic occupancy. This weighting enforces dynamic temporal aggregation only in dynamic occupied regions, ignoring flow predictions in free space and inside static objects. With this dynamic aggregation, the geometric losses applied to the aggregated dynamic field $\bar{\phi}^{d}_t$ implicitly train the flow head by encouraging accurate flow predictions in dynamic regions. The total blended aggregated SDF $\bar{\phi}^{b}_t$ is then obtained using Eq. \ref{eq:sdfblending}.

\textbf{Similarity flow loss.} 
Jointly learning geometry and scene flow without using labels or pretrained optical flow models is challenging due to the degrees of freedom introduced by our flow-driven dynamic warping. To address this, we design a similarity flow loss $\Loss_{sim}$ that explicitly supervises the flow head using pseudo-labels derived from the BEV feature similarities (see Fig. \ref{fig:bevfeatssim}). During training, we copy and stop the gradients of the dynamic BEV features at current $\beta^{d}_{t}$ and previous $\beta^{d}_{t-1}$ timesteps, aligned by ego motion. For each grid cell $(i,j)$ in $\beta^{d}_{t}$, we compute the cosine similarities with its neighboring cells from $\beta^{d}_{t-1}$ within an $N\!\times\!N$ search window:

\begin{equation}
\begin{aligned}
    s_{t}(i, j, \Delta i, \Delta j) = 
\cos\left( \beta^{d}_{t}(i, j), \, \beta^{d}_{t-1}(i + \Delta i, j + \Delta j) \right), \\
- \left\lfloor \tfrac{N}{2} \right\rfloor \leq \Delta i, \Delta j \leq \left\lfloor \tfrac{N}{2} \right\rfloor, \quad \cos(\mathbf{x}, \mathbf{y}) = \tfrac{\mathbf{x}^\top \mathbf{y}}{\|\mathbf{x}\| \, \|\mathbf{y}\|}
\end{aligned}
\end{equation}

We then take the displacement of the most similar neighbor: 

\begin{equation}
s^{\text{max}}_{t}(i,j) = \underset{(\Delta i, \Delta j) \in \mathcal{N}}{\mathrm{argmax}} s_t(i, j, \Delta i, \Delta j)
\end{equation}

We multiply $s^{\text{max}}_{t}$ by the cell size to obtain flow in metric scale and broadcast it along the Z axis to form the backward pseudo-labels $\mathbf{f}^{s}_{t^-}$. The same procedure with $\beta^{d}_{t}$ and $\beta^{d}_{t+1}$ produces the forward pseudo-labels $\mathbf{f}^{s}_{t^+}$. The flow predictions $\mathbf{f}_{t^-}$, $\mathbf{f}_{t^+}$ are regressed to the pseudo-labels $\mathbf{f}^s_{t^-}$, $\mathbf{f}^s_{t^+}$ using an L1 loss. However, geometric inconsistencies between adjacent frames' predictions can introduce noise in $\mathbf{f}^{s}$, especially at early training. To mitigate this issue, we weight down the loss contribution at regions where $\mathbf{f}^{s}_{t^-}$ and $\mathbf{f}^{s}_{t^+}$ disagree using an exponential decay. The forward-backward consistency weight $\gamma_{s}$ at a given 3D cell $(i,j,k)$ is computed as:

\begin{equation}
\gamma_{s}(i,j,k) = \exp\Big(-\tau_{s} \big\| \mathbf{f}^{s}_{t^-}(i,j,k) + \mathbf{f}^{s}_{t^+}(i,j,k) \big\|_2\Big)
\end{equation}

where $\tau_{s}$ is a constant scalar that controls the decay rate of $\gamma_{s}$. As we do in the temporal aggregation, we restrict this loss to dynamic occupied regions by multiplying it by the dynamic occupancy $\Phi^{d}_{a}$. The final similarity flow loss is therefore:

\begin{equation}
\Loss_{sim} = \Phi^{d}_{a} \cdot \gamma_{s} \Big(\big|\mathbf{f}_{t^-} \!- \mathbf{f}^{s}_{t^-}\big| + \big|\mathbf{f}_{t^+} \!- \mathbf{f}^{s}_{t^+}\big| \Big)
\end{equation}

\textbf{Ray-based supervision.} We supervise our model by casting both camera and LiDAR rays. We sample $M$ points $\mathbf{P} = \{\mathbf{p}_{m}|m=1,...,M\}$ at uniform distances from the sensor's origin until its intersection with the representation boundaries. We use NeuS' \cite{NeuS} formulation to obtain the probability of a ray terminating between $p_{m}$ and $p_{m+1}$:

\begin{equation}
\label{alphaneus}
\alpha_{m} = \max\Bigg[\frac{\Phi_{a}\big(\phi (\mathbf{p}_{m})\big) - \Phi_{a}\big(\phi(\mathbf{p}_{m+1})\big)}{\Phi_{a}\big(\phi (\mathbf{p}_{m})\big) }, 0 \Bigg]
\end{equation}

where $\Phi_{a}(x) = (1+e^{-ax})^{-1}$ is the sigmoid function with $a$ being a learnable parameter that controls its sharpness. The accumulated transmittance is therefore calculated as $T_{m} = \prod_{j=1}^{m-1} (1-\alpha_{j}) $ and the probability of the ray ending at $p_{m}$ after emission is $w_{m} = T_{m}\alpha_{m}$. The rendering process is an integration of per-point attributes over the probability distribution $\textbf{w} = \{w_{m}|m=1,...,M\}$, and the rendered color $\textbf{c}_{r}$ and depth $d_{r}$ of the ray can be calculated as:

\begin{equation}
\textbf{c}_{r} = \sum_{m=1}^{M} w_{m} \mathbf{c}_{m}, \quad {d}_{r} = \sum_{m=1}^{M} w_{m} d_{m}
\end{equation}

with $\mathbf{c}_{m}$ and $d_{m}$ being the color and depth of the ray's $m$-th point. 

We supervise our model with camera rays using a photometric loss $\Loss_{photo}$. For depth supervision we use SelfOcc's \cite{SelfOcc} depth loss $\Loss_{dep}$, which extends the receptive field of the standard reprojection loss $\Loss_{rpj}$ (common in self-supervised depth estimation) across the full epipolar line. It also incorporates Monodepth's \cite{monodepth2} min-loss and auto-masking to handle occlusions, moving objects and low texture regions. In addition, we include an RGB loss $\Loss_{rgb}$ that regresses the rendered colors $\mathbf{c}_{r}$ with the target image ones $\mathbf{I}_{t}$ using a combination of L1 and D-SSIM losses, following Monodepth \cite{monodepth2}. The total photometric loss is: $\Loss_{photo} = \lambda_{dep}\Loss_{dep} + \lambda_{rgb}\Loss_{rgb}$.

\begin{figure*}[t]
\centering
\includegraphics[width=0.99\textwidth]{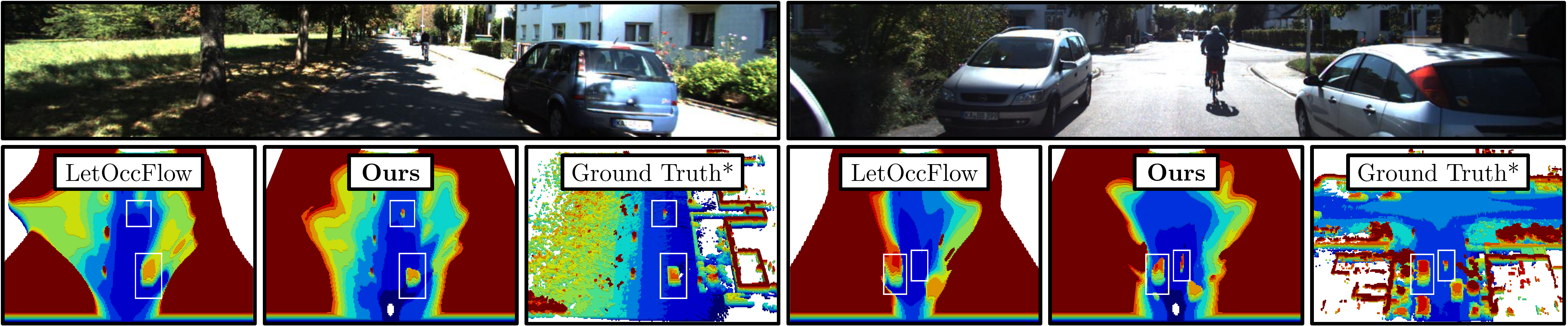}
\caption{\textbf{3D occupancy comparison on SemanticKITTI \cite{SemanticKITTI}}. We draw white boxes on areas with notable changes. Our model predicts better occupancy for small dynamic objects, and infers better geometries in occluded regions such as behind cars.}
\label{fig:qualitativekittigeo}
\end{figure*} 

We also use LiDAR rays to supervise the geometry predictions by employing an L2 range loss $\Loss_{r}$ between each ray's predicted depth and its measured range. At each iteration, half of the sampled rays are static and half are dynamic. Static rays are sampled over the aggregated static SDF $\bar{\phi}^{s}_t$ and can either belong to the current timestep $t$ or to neighboring ones $\{\, t\!\pm\!k \mid k \in \{0, 1, \dots, K\!-\!1\} \,\}$, leveraging the stationary nature of $\bar{\phi}^{s}_t$. Supervising the static field with rays from multiple timesteps allows learning geometry at occluded areas, and reduces static occupancy in dynamic regions. Dynamic rays are sampled over the aggregated dynamic SDF $\bar{\phi}^{d}_t$ but only those from the current timestep are used to avoid potential temporal inconsistencies from moving objects. To counter the excessive dynamic field sparsity caused by per-field regularization and sparsity losses, we introduce a dynamic density loss $\Loss_{den}(\mathbf{p}^{d}) = \max(\phi^{d}_{t}(\mathbf{p}^{d}), 0)$ which penalizes empty dynamic predictions at the points $\mathbf{p}^{d}$ where a dynamic LiDAR ray ends. The total LiDAR loss is: $\Loss_{lidar} = \lambda_{r}\Loss_{r} +  \lambda_{den}\Loss_{den}$.

\textbf{Regularization losses.} 
We apply eikonal and hessian regularization \cite{SelfOcc} to both static and dynamic fields $\Loss^{s}_{e}, \Loss^{s}_{H}, \Loss^{d}_{e}, \Loss^{d}_{H}$, and a hessian loss $\Loss^{\mathbf{f}}_{H}$ on the flow field to enhance spatial smoothness. To promote static over dynamic occupancy in background regions, we apply a sparsity loss $\Loss^{d}_{s} = \max(-\phi^{d}_{t}, 0)$ at uniformly sampled points in the dynamic field. The total regularization loss $\Loss_{reg}$ is:

\begin{equation}
\label{eq:lossreg}
\Loss_{reg} = \lambda^{s}_{e}\Loss^{s}_{e} + \lambda^{d}_{e}\Loss^{d}_{e} + \lambda^{s}_{H}\Loss^{s}_{H} + \lambda^{d}_{H}\Loss^{d}_{H} + \lambda^{\mathbf{f}}_{H}\Loss^{\mathbf{f}}_{H} + \lambda^{d}_{s}\Loss^{d}_{s}
\end{equation}

Finally, the total loss combines all objectives:

\begin{equation}
\Loss = \lambda_{sim}\Loss_{sim} + \Loss_{photo} + \Loss_{lidar} + \Loss_{reg} 
\label{eq:losstotal}
\end{equation}

\section{Results}

\textbf{Data.} We evaluate our approach experimentally on SemanticKITTI, KITTI-MOT, and nuScenes. SemanticKITTI \cite{SemanticKITTI}, derived from KITTI Vision Benchmark, provides LiDAR scans synchronized with two forward-facing stereo RGB cameras, from which we only use the left one (cam2). We follow the official splits using sequence 08 for validation and the remaining 00-10 for train. KITTI-MOT \cite{KITTIMOT} is based on the KITTI Multi-Object Tracking evaluation 2012 benchmark, consisting of 21 training and 29 testing sequences. We match the setup of \cite{LetOccFlow} and select training sequences 0000-0011, 0014-0016, 0018-0020 for train, and testing sequences 0000-0014 for validation. NuScenes \cite{nuScenes} consists of 1000 sequences of 20 seconds duration with RGB images from 6 surround cameras with 360° FOV collected at 12 Hz, from which a subset of keyframes at 2 Hz contain LiDAR scans synchronized with the cameras. We follow the official splits of 700 training and 150 validation scenes. 

\begin{table}[h]
\centering
\caption{\textbf{3D occupancy prediction on SemanticKITTI \cite{SemanticKITTI}}. For OccNeRF \cite{OccNeRF}, we use the results from LetOccFlow's \cite{LetOccFlow} adaptation to SemanticKITTI \cite{SemanticKITTI}.}
\begin{tabular}{c|cccc}
\toprule
\multirow{2}{*}{\textbf{Method}} & \multicolumn{4}{c}{\textbf{3D Occupancy}} \\
& \multicolumn{3}{c}{RayIoU\textsubscript{1m,2m,4m}$\uparrow$} & RayIoU$\uparrow$ \\
\midrule
MonoScene \cite{MonoScene} & 26.00 & 36.50 & 49.09 & 37.19 \\
SceneRF \cite{SceneRF} & 20.65 & 35.74 & 56.35 & 37.58  \\
OccNeRF \cite{OccNeRF} & 23.22 & 40.17 & 62.57 & 41.99  \\
SelfOcc \cite{SelfOcc} & 23.13 & 35.85 & 50.31 & 36.27  \\
LetOccFlow \cite{LetOccFlow} & 28.62 & 45.69 & 66.95 & 47.06  \\
\midrule
Ours \textit{(w/o T.A.)} & 27.58 & 44.03 & 65.81 & 45.81  \\
\textbf{Ours} & \textbf{29.57} & \textbf{48.21} & \textbf{72.83} & \textbf{50.20}  \\
\bottomrule
\end{tabular}
\label{tab:occsemkitti}
\end{table}

\textbf{Implementation details.} 
We use a BEV \cite{BEVFormer} representation uniformly divided to cover the cuboid volume of [51.2, 51.2, 6.4] meters in front of the ego car, with 0.2 m resolution in SemanticKITTI \cite{SemanticKITTI} and KITTI-MOT \cite{KITTIMOT}, and [80, 80, 6.4] meters around the ego with 0.4 m resolution in nuScenes \cite{nuScenes}.

The static and dynamic heads are both 2-layer MLPs with Softplus activations that predict SDF and RGB values at every height for each BEV cell. The flow head is a 2D multi-scale U-Net \cite{UNet} with an encoder-decoder structure and skip connections for multi-scale feature fusion. It uses 3$\times$3 convolutions with stride 2 for downsampling to $\tfrac{1}{2}$ and $\tfrac{1}{4}$ resolution, and bilinear interpolation for upsampling, followed by linear layers to fuse features at every scale. Each convolution is followed by batch normalization and ReLU. The final linear layer outputs backward and forward flow vectors at each 3D grid cell.

\begin{figure*}[t]
\centering
\includegraphics[width=0.99\textwidth]{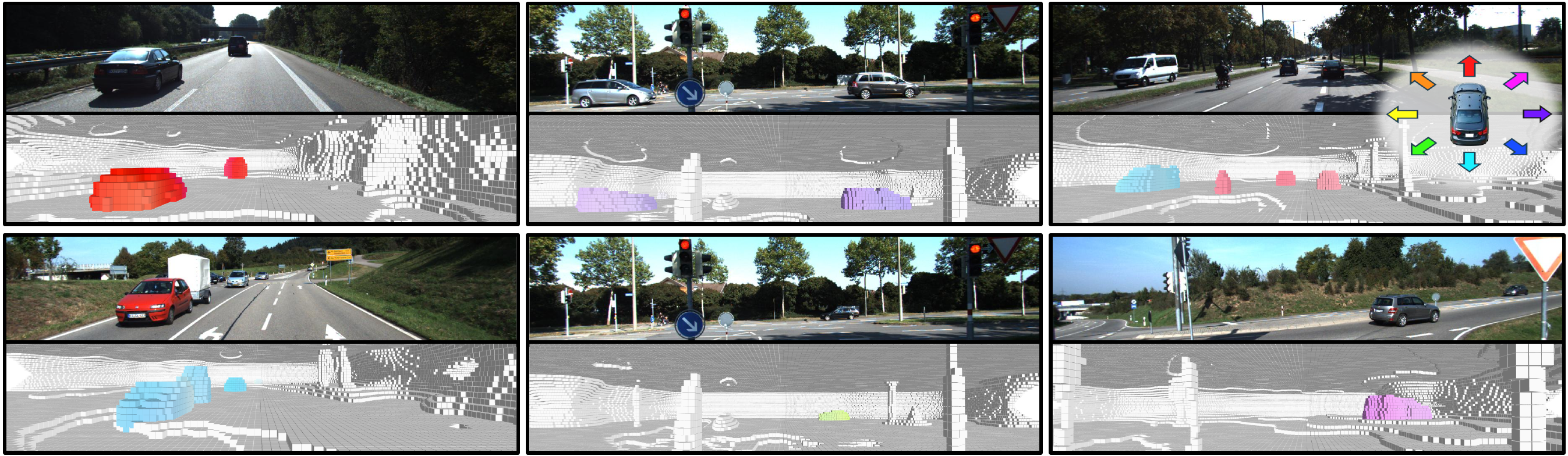}
\caption{\textbf{3D occupancy flow results on KITTI-MOT \cite{KITTIMOT}}. Flow legend displayed at the top right corner.} 
\label{fig:qualitativekittiflow}
\end{figure*}

\begin{table*}[h]
\centering
\caption{\textbf{3D occupancy flow prediction on KITTI-MOT \cite{KITTIMOT}.} \textbf{Ours}-\textit{SemKITTI} is trained on SemanticKITTI \cite{SemanticKITTI} and evaluated on KITTI-MOT \cite{KITTIMOT} without fine-tuning (not directly comparable, shown to illustrate generalization).}
\begin{tabular}{c|c|ccccccc}
\toprule
\multirow{2}{*}{\textbf{Method}} & \multirow{2}{*}{\textbf{Supervision}} & \multicolumn{7}{c}{\textbf{3D Occupancy \& Occupancy Flow}} \\
& & DE$\downarrow$ & EPE$\downarrow$ & $\text{DE}_{\text{FG}}\!\downarrow$ & $\text{EPE}_{\text{FG}}\!\downarrow$ & $\text{D}_{5\%}\!\downarrow$ & $\text{Fl}_{10\%}\!\downarrow$ & $\text{SF}_{10\%}\!\downarrow$ \\
\midrule
OccNeRF-C\* \cite{OccNeRF} & C+O & 3.547 & 6.986 & 7.976 & 10.445 & 0.204 & 0.272 & 0.306 \\
LetOccFlow-C\* \cite{LetOccFlow} & C+O & 3.431 & 3.529 & 6.429 & 7.635 & 0.215 & 0.084 & 0.118 \\
\midrule
OccNeRF-L\* \cite{OccNeRF} & C+L+O & 2.667 & 8.131 & 4.040 & 11.284 & 0.137 & 0.309 & 0.326 \\
LetOccFlow-L\* \cite{LetOccFlow} & C+L+O & 2.610 & \textbf{3.488} & 3.106 & \textbf{7.510} & 0.154 & \textbf{0.083} & \textbf{0.105} \\
\textbf{Ours}  & C+L & \textbf{1.907} & 6.788 & \textbf{2.435} & 11.638 & \textbf{0.112} & 0.171 & 0.182 \\
\midrule
\textbf{Ours}-\textit{SemKITTI} & C+L & 1.655 & 5.685 & 2.259 & 11.130 & 0.087 & 0.148 & 0.158 \\
\bottomrule
\end{tabular}
\label{tab:flowkittimot}
\end{table*}

The input image resolution is [370, 1216] pixels for SemanticKITTI and KITTI-MOT, and [384, 800] for nuScenes. We set $\tau \!=\!2.0$, $\lambda_{ag}\!=\!0.5$, $\lambda_{r}\!=\!10.0$, $\tau_{s}\!=\!0.75$, $K\!=\!20$, $\lambda_{sim}\!=\!5.0$, $\lambda_{dep}\!=\!1.0$, $\lambda_{rgb}\!=\!0.1$, $\lambda^{s}_{e}\!=\!\lambda^{d}_{e}\!=\!0.1$, $\lambda^{s}_{H}\!=\!\lambda^{d}_{H}\!=\!0.1$, $\lambda^{\mathbf{f}}_{H}\!=\!0.02$ and $\lambda_{den}\!=\!0.01$. We use $N\!=\!35$ to ensure that $\Loss_{sim}$ captures inter-frame displacements up to highway speeds while keeping GPU memory overhead low. Its memory footprint can be further reduced by using $\texttt{stride}\!>\!1$ with bilinear interpolation or by lowering $N$ for higher-framerate data or coarser BEV grids. We set $\lambda^{d}_{s}\!=\!0.01$ for SemanticKITTI and KITTI-MOT, and $\lambda^{d}_{s}\!=\!0.1$ for nuScenes. We jointly optimize our multi-objective losses (photometric, LiDAR, similarity flow and regularization) ensuring stable convergence and balanced magnitudes. To set the loss weights, we first normalized each term to a similar scale and then fine-tuned the weights empirically, observing training stability across multiple runs without exploding gradients or NaNs. We train our models for 24 epochs using a total batch size of 4 on 4 NVIDIA Tesla V100 GPUs (32 GB each).

The dynamic masks used for ray pseudo-label generation are obtained using Grounded-SAM \cite{GroundedSAM}. This choice ensures fair comparison with LetOccFlow \cite{LetOccFlow} and highlights that our performance gains are attributable to the proposed methodology and not to the foundation model's performance. We obtain dynamic masks by querying Grounded-SAM with the following prompts: \textit{bicycle}, \textit{bus}, \textit{car}, \textit{motorcycle}, \textit{person}, \textit{train}, \textit{truck}, and \textit{van}. We keep the masks with confidence $\ge 0.5$ to label dynamic rays. Regions with no mask of confidence $\ge 0.3$ are labeled as static, and the remaining uncertain areas are discarded. The 3D voxel clusters are computed using the 3D connected components algorithm \cite{Allegretti2019GPU} ($3\!\times\!3\!\times\!3$ kernel) on voxelized LiDAR points at resolution 0.2 for SemanticKITTI and KITTI-MOT, and 0.4 for nuScenes.

\begin{figure*}[t]
\centering
\includegraphics[width=0.99\textwidth]{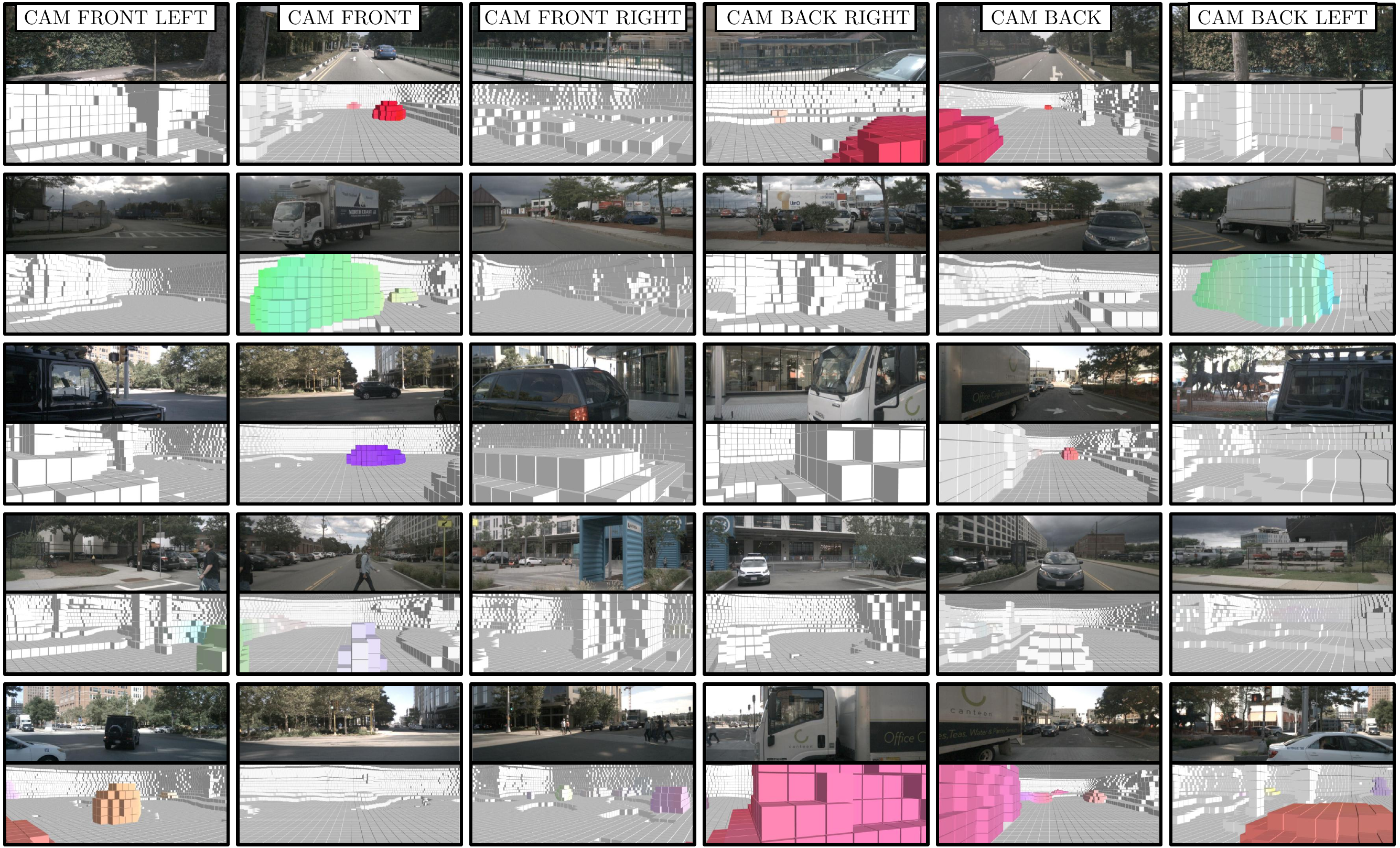}
\caption{\textbf{3D occupancy flow results on nuScenes \cite{nuScenes}}. Flow colors are the same as in KITTI-MOT.} 
\label{fig:qualitativenuScenesoccflow}
\end{figure*}

\textbf{Experiments on SemanticKITTI.} We first show the effectiveness of our temporal aggregation mechanism on learning accurate and temporally consistent scene geometry. Table \ref{tab:occsemkitti} shows the 3D occupancy results of our model with and without temporal aggregation. We use the RayIoU metric proposed by \cite{RayIoU}, consisting of uniformly sampling LiDAR-like rays originating at the current and the next 8 ego poses, and comparing the depths between predicted and ground-truth occupancy on the KITTI's Semantic Scene Completion (SSC) benchmark \cite{SemanticKITTI}. The rays whose depth error is lower than a threshold are classified as \textit{true positives}. This strategy enables evaluation on occluded areas (e.g., behind a car or tree), while ignoring never-observed regions (e.g., behind buildings). We report the RayIoU with 1, 2 and 4 meters thresholds and their average. Our full model improves RayIoU by +4.39\% over the version without temporal aggregation and similarity flow. We also report and compare the RayIoU with previous camera-based occupancy models, improving the LetOccFlow results by +3.14\%. Fig. \ref{fig:qualitativekittigeo} visualizes the predicted 3D occupancy from our model and LetOccFlow's official checkpoint, along with a refined ground-truth used only for visualization (combining the SSC official labels with SemanticKITTI's LiDAR motion annotations to remove artifacts). Our model predicts significantly better occupancy in small and rare dynamic objects (e.g., bikers) and in occluded regions (e.g., behind cars).

\textbf{Experiments on KITTI-MOT.} Due to the absence of occupancy flow annotations, we evaluate the occupancy flow predictions by following the same strategy as \cite{LetOccFlow}, which uses the metrics from KITTI's scene flow benchmark \cite{KITTIMOT}. For each evaluation image, we generate disparity pseudo ground-truth using the LiDAR points' projected depths, and obtain optical flow pseudo-labels by querying the pretrained optical flow model used in \cite{LetOccFlow}. We render depth and flow at each projected point, convert them to disparity and optical flow, and compare them with the pseudo-labels. In Table \ref{tab:flowkittimot}, we report the end-point error of disparity (DE) and optical flow (EPE). $\text{D}_{\text{5\%}}$ represents the percentage of disparity outliers with EPE $<$ 4 pixels or 5\% of the disparity pseudo ground-truth. $\text{Fl}_\text{10\%}$ is the percentage of optical flow outliers with EPE $<$ 8 pixels or 10\% of the flow pseudo ground-truth. $\text{SF}_{\text{10\%}}$ is the percentage of scene flow outliers (outliers in either $\text{D}_{\text{10\%}}$ or $\text{Fl}_\text{10\%}$). We also report the foreground disparity error ($\text{DE}_{\text{FG}}$) and optical flow error ($\text{EPE}_{\text{FG}}$) computed on pixels within dynamic masks from \cite{GroundedSAM}.

Our model achieves the highest performance across all depth estimation metrics and provides competitive optical flow results, despite not using 2D optical flow supervision. Unlike the rest of the methods which directly regress these optical flow pseudo-labels, our model learns dynamic objects' motion implicitly, making these results particularly significant given that evaluation occurs in the same image domain where the pseudo-labels were generated. To emphasize the generalization ability of our model to unseen driving scenes, we include in the last row of Table \ref{tab:flowkittimot} (\textbf{Ours}-\textit{SemKITTI}) the performance of our model trained on SemanticKITTI and directly evaluated on KITTI-MOT, without any fine-tuning. The results show slightly higher scores across all metrics, likely due to the larger amount of training samples in SemanticKITTI. This suggests good generalization ability of our model to new scenes with equivalent setups. Fig. \ref{fig:qualitativekittiflow} shows qualitative examples of our model's occupancy and flow predictions on KITTI-MOT.

\textbf{Experiments on nuScenes.} We evaluate the 3D occupancy flow performance of our model in nuScenes using the occupancy flow labels from CVPR's 2024 Autonomous Grand Challenge \cite{openoccv2} and report the results in Table \ref{tab:occflownuScenes}. We compute the RayIoU \cite{RayIoU} and mean Average Velocity Error (mAVE) metrics as in OccNet \cite{openoccv2}. The mAVE is defined in meters per second and evaluated only in occupied voxels of dynamic classes: \textit{car, truck, trailer, bus, construction vehicle, bicycle, motorcycle, pedestrian}. The mAVE is computed by sampling the 3D flow prediction at the ray's endpoint for every \textit{true positive} from RayIoU's evaluation using a 2-meter threshold.

Our model improves the RayIoU by +1.73\% compared to OccNet and +0.91\% compared to LetOccFlow trained with additional LiDAR supervision. Regarding flow estimation, our model reduces the mAVE by 7.7\% with respect to LetOccFlow, establishing a new state-of-the-art for 3D occupancy flow prediction on nuScenes. Fig. \ref{fig:qualitativenuScenesoccflow} shows qualitative examples of our model's occupancy and flow predictions on nuScenes.

\begin{table}[h]
\centering
\caption{\textbf{3D occupancy flow prediction on nuScenes \cite{nuScenes}.} OccNet is trained with 3D labels. RenderOcc, OccNeRF-L, LetOccFlow-L and \textbf{ours} use LiDAR supervision. All methods except \textbf{ours} (and OccNet) use optical flow supervision.}
\begin{tabular}{c|cccc|c}
\toprule
\multirow{2}{*}{\textbf{Method}} & \multicolumn{5}{c}{\textbf{3D Occupancy \& Occupancy Flow}} \\
& \multicolumn{3}{c}{RayIoU\textsubscript{1m,2m,4m}$\uparrow$} & RayIoU$\uparrow$ & mAVE$\downarrow$ \\
\midrule
OccNet \cite{openoccv2} & \textbf{29.28} & 39.68 & 50.02 & 39.66 & 1.61  \\
\midrule
OccNeRF-C\* \cite{OccNeRF} & 9.93 & 19.06 & 35.84 & 21.61 & 1.53  \\
LetOccFlow-C\* \cite{LetOccFlow}  & 17.49 & 28.52 & 44.33 & 30.12 & 1.42  \\
\midrule
RenderOcc\* \cite{RenderOcc} & 20.27 & 32.68 & 49.92 & 36.67 & 1.63  \\
OccNeRF-L\* \cite{OccNeRF} & 16.62 & 29.25 & 49.17 & 31.68 & 1.59  \\
LetOccFlow-L\* \cite{LetOccFlow} & 25.49 & 39.66 & 56.30 & 40.48 & 1.45  \\
\textbf{Ours}  & 25.07 & \textbf{40.46} & \textbf{58.64} & \textbf{41.39} & \textbf{1.31}  \\
\bottomrule
\end{tabular}
\label{tab:occflownuScenes}
\end{table}

\textbf{Efficiency evaluation.} We compare the complexity of our model and LetOccFlow in terms of number of parameters, frames per second (FPS), and floating-point operations (FLOPs) to highlight the lightweight advantage of our method. Our model avoids LetOccFlow's costly dense 3D convolutions, uses notably lower channel dimensions, and operates on a lightweight BEV representation instead of TPV. As shown in Table \ref{tab:efficiencysemkitti}, these design choices lead to significantly reduced compute complexity, lower memory requirements, and faster inference times on a single V100.

\begin{table}[h]
\centering
\caption{\textbf{Model complexity and inference efficiency on SemanticKITTI \cite{SemanticKITTI}.} The values for LetOccFlow \cite{LetOccFlow} are obtained experimentally using their official checkpoint.}
\begin{tabular}{c|ccc}
\toprule
Method & Params (M)$\downarrow$ & FPS$\uparrow$ (V100)  & FLOPs (G)$\downarrow$ \\
\midrule
LetOccFlow \cite{LetOccFlow} & 253.3 & 1.04 & 3202 \\ 
\textbf{Ours} & \textbf{32.4} & \textbf{3.78} & \textbf{405} \\ 
\bottomrule
\end{tabular}
\label{tab:efficiencysemkitti}
\end{table}

\begin{table}[h]
\centering
\caption{\textbf{Ablation study on KITTI-MOT \cite{KITTIMOT}.} Individual contributions of static and dynamic temporal aggregation.}
\begin{tabular}{c|ccccc}
\toprule
Method & $\text{DE}\!\downarrow$ & $\text{EPE}\!\downarrow$ & $\text{D}_{\text{5\%}}\!\downarrow$ & $\text{Fl}_{\text{10\%}}\!\downarrow$ & $\text{SF}_{\text{10\%}}\!\downarrow$ \\
\midrule
Ours \textit{(w/o T.A.)} & 1.948 & 7.074 & 0.116 & 0.179 & 0.191\\
Ours \textit{(w/o dyn. T.A.)} & 1.909 & 6.861 & 0.114 & 0.177 & 0.188\\
\textbf{Ours} & \textbf{1.907} & \textbf{6.788} & \textbf{0.112} & \textbf{0.171} & \textbf{0.182}\\
\bottomrule
\end{tabular}
\label{tab:ablation}
\end{table}

\textbf{Ablation study.} To assess the contribution of each main component of our method to flow learning, we conduct experiments where we disable individual components and evaluate the resulting performance. From our earliest experiments, we observed that removing $\Loss_{sim}$ prevents the model from learning flow altogether, yielding meaningless outputs. We also found that modeling static and dynamic geometries as a single SDF makes the temporal aggregation and $\Loss_{sim}$ computation ill-defined, especially near contact regions between static and dynamic objects (e.g., between a car and the road). In contrast, removing static and/or dynamic temporal aggregation degrades performance, but does not prevent convergence. Table \ref{tab:ablation} reports depth and flow metrics for these variants, including a version with no temporal aggregation (\textit{Ours w/o T.A.}), and another without the dynamic one (\textit{Ours w/o dyn. T.A.}). The results indicate that both temporal aggregation modules are beneficial but not crucial, while scene disentanglement and $\Loss_{sim}$ are critical for jointly learning geometry and motion.

\textbf{Failure cases.} The lack of fine-grained occupancy and flow labels complicates quantitative evaluation in challenging cases such as non-rigid motions or thin dynamic objects. However, we observed qualitatively that our model struggles to estimate motion for nearby objects moving in opposite directions, likely due to the flow smoothness constraint imposed by $\Loss^{\mathbf{f}}_{H}$. Very large motions in KITTI's German highways were also not fully captured, possibly due to the limited spatial context available to the $\Loss_{sim}$ and flow head. Moreover, moving pedestrians were mostly predicted as rigid moving blocks.

\section{Conclusion}

This work presents an effective approach for learning 3D occupancy flow without relying on costly human-produced annotations or pretrained optical flow models. Our method jointly learns geometry and motion by disentangling the scene into separate static and dynamic SDFs, enforcing temporal invariance in static geometry, and ensuring spatio-temporal consistency for dynamic elements. We further introduce a self-supervised flow cue derived from features' cosine similarities between consecutive frames to facilitate convergence. By removing the dependence on labels or external flow supervision, our approach represents a step towards end-to-end self-supervised 3D occupancy flow prediction.

\end{document}